# Multimodal Strain Sensing System for Shape Recognition of Tensegrity Structures by Combining Traditional Regression and Deep Learning Approaches

Zebing Mao[1], Ryota Kobayashi[2], Hiroyuki Nabae[2], and Koichi Suzumori[2]

*Abstract*—A tensegrity-based system is a promising approach for dynamic exploration of uneven, unpredictable, and confined environments. However, implementing such systems presents challenges in state recognition. In this study, we introduce a 6-strut tensegrity structure integrated with 24 multimodal strain sensors, employing a deep learning model to achieve smart tensegrity. By using conductive flexible tendons and leveraging a long short-term memory (LSTM) model, the system accomplishes self-shape reconstruction without the need for external sensors. The sensors operate in two modes, and we applied both a curve fitting model and an LSTM model to establish the relationship between length change and resistance change in the sensors. Our key findings demonstrate that the intelligent tensegrity system can accurately self-detect and adapt its shape. Furthermore, a human pressing process allows users to monitor and understand the tensegrity's shape changes based on the integrated models. This intelligent tensegrity-based system with self-sensing tendons showcases significant potential for future exploration, making it a versatile tool for real-world applications.

## I. INTRODUCTION

The concept of using tensegrity structures in extreme environmental exploration is an innovative approach that offers several advantages due to the unique properties of tensegrity systems [1]. One famous example is the "Super Ball Bot" developed by NASA (National Aeronautics and Space Administration) [2][3], which can be used for space exploration. Tensegrity structures are composed of solid compression components (rods/struts) connected by tension elements (cables/strings). These assemblies create structures that are both light and sturdy, maintaining a balance between the forces of compression and tension.

Over recent years, A myriad of tensegrity robot designs capable of movement have emerged, featuring diverse actuation methods and mobility techniques [4][5][6]. These robots are powered by various actuation systems such as electric motors [7], pneumatic actuators [8], shape memory alloys [9][10], and dielectric elastomers [11] etc. Typical mobility strategies employed by these robots include methods like swimming [12], rolling [13], climbing [14], jumping [15], twisting [16] etc. Previous efforts have yielded numerous effective designs and movement strategies for tensegrity structures, enabling them to navigate terrains smoothly in a sensor-free, open-loop pattern. However, the

Zebing Mao is with the Faculty of Engineering, Yamaguchi University, Yamaguchi 755-8611, Japan (mao.z.aa@yamaguchi-u.ac.jp).
Ryota Kobayashi, Hiroyuki Nabae, Koichi Suzumori are with the Department of Mechanical Engineering, Tokyo Institute of Technology, Tokyo 152-8552, Japan.

challenging conditions of severe exploration can disrupt their movement and alter their shapes due to the rough terrain. Consequently, the need for the intelligent systems including state reconstruction of tensegrity structures has become critical, especially when we consider their practical applications.

Unfortunately, few literatures addressing this aspect of tensegrity have been previously reported, although a large number of soft and flexible sensors have been developed [17][18][19]. Jonathan Bruce et al., measured the varying lengths of the tensile elements connecting the rods to precisely calculate the spatial distance between two nodes in the design, but there are limited sensors capable of measuring such significant deformations [20]. Ken Caluwaerts et al., developed a technique that utilizes the unscented Kalman filter (UKF), integrating inertial measurements, ultra-wideband time-of-flight ranging data, and information about actuator states [21]. However, relying on external sensors, markers, or designated base stations to enhance system resolution is cumbersome and limit the tensegrity's flexibility. Alternatively, Joran W. Booth et al., showcased a method for real-time state reconstruction of a tensegrity robot, employing robotic skins equipped with pneumatic actuators and embedded strain sensors [22]. In their physical tensegrity robot, the distance between the nodes differs from the active length of the sensors because of the way the sensors are attached to the structure. Wen-Yung Li et al., achieved shape recognition by employing a tensegrity structure equipped with a soft sensor, using a recurrent neural network method [23]. But they collected the data firstly and then reconstructed, which is not real-time process. J. Kimber et al., developed six-axis accelerometer to record the tensegrity's vibrations but with the ability of accomplishing partial state information [24]. Among them, these physics-based models are excellent for designing a new tensegrity system and balancing the required tension forces within the system [25][26]. However, these proposed physics-based models are not well suited for state reconstruction since they rely on known spring forces for system components, tension on the cables, and torques on the bars, which are difficult to measure. Furthermore, the root mean squared error (RMSE) in shape recognition is notably high, with values exceeding 30 mm for individual nodes and 40 mm for the entire system. This level of error can lead to significant inaccuracies and potential mistakes in applications requiring precise measurements, such as in confined spaces.

Therefore, we focus on the six-strut tensegrity, characterized by its dormant state forming an irregular icosahedron, with 8 equilateral and 12 isosceles triangles forming its surface. We choose the type of tensegrity because the six-strut tensegrity structure's stability, efficiency, and adaptability highlight its morphological significance in extreme environmental exploration, offering a robust and practical solution for challenging conditions. After designing and manufacturing a 6-strut tensegrity, we replace the tendons with soft multimodal sensors. We then use the LSTM (Long Short-Term Memory) method to extract models of stretching properties and a curve-fitting equation to estimate the bending characteristics. Finally, we achieve shape recognition of the tensegrity.

## II. Theory And Method

This section outlines the essential requirements for employing a method to replicate tensegrity in a virtual environment. The tensegrity structure in this study consists of six wood-made struts and twenty-four rubber tendons, forming the shape of an icosahedron (Fig. 1). The model comprises multiple independent rigid rods and multiple independent flexible lines, where each rigid body has six degrees of freedom

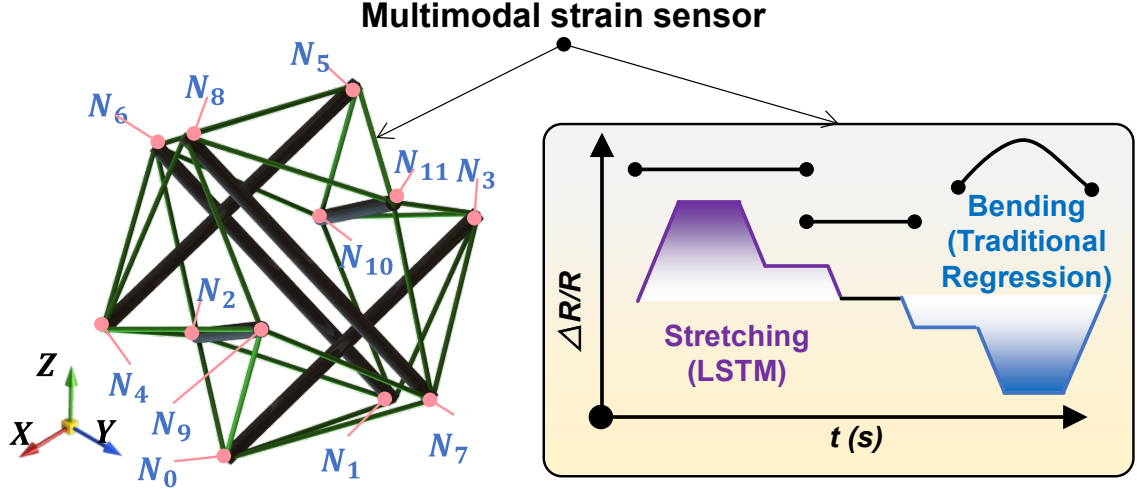

Fig. 1 Six-strut tensegrity structure with labeled nodes and methods

Our strain sensor approach takes direct measurements that can be correlated with distances between nodes and enables state reconstruction of the tensegrity robot without any knowledge of the forces in the system. The method contains two primary sections: (i) Data acquisition module, (ii) Shape recognition (Fig. 2). For the data acquisition module (Fig. 2A), we used two Arduino Mega sets, which can power 24 measured sensors with their 5V power supply and ground. To calculate the multimodal strain sensors in real time, we used 24 known resistors (5.8 MΩ) connected in a parallel pattern. The state of the six-strut tensegrity can be described using a mathematical model: a matrix of node positions in a locally defined coordinate system, which gives the shape of the system. The sensor readings are also fed to an algorithm that estimates the node positions ($N$) by solving 27 geometrical equations. By this method, we can obtain the values of 24 sensors in the computer. Next, we can reconstruct the tensegrity based on the following algorithms (Fig. 2**B**). Since the sensor has two states: bending rate and stretching state, we will build two models using polynomial fitting curves and machine learning methods. More specifically, two sets of Arduino Mega will provide the resistive values of 24 sensors ($R$) to the trained models ($f_3$) or fitting equation ($f_4$) to calculate the strain ($\varepsilon$):

$$R = [R_0, R_2 \dots R_{24}]^T \tag{1}$$

$$\varepsilon = [\varepsilon_0, \varepsilon_2 \dots \varepsilon_{24}] \tag{2}$$

Then, the relationship can be defined as:

$$\varepsilon = \begin{cases} f_3(R) & \text{streching state} \\ f_4(R) & \text{bending stare} \end{cases} \tag{3}$$

Therefore, the lengths (L) of all tendons (24 sensors) can be given by the formula:

$$L = [L_0, L_2 \dots L_{23}]^T = (1 + \varepsilon)[L_{0o}, L_{2o} \dots L_{23o}]^T \tag{4}$$

where $L_{1o}, L_{2o} \dots L_{24o}$ are the original lengths of the 24 tendons. Additionally, we prepared the coordinates to model the shape of the tensegrity, in which the positions of three nodes ($N_0$ ($x_0, y_0, z_0$), $N_1$ ($x_1, y_1, z_1$), $N_2$ ($x_2, y_2, z_2$)) at the bottom are defined based on their own origins. After setting the origin, the positions of the other nodes can be determined using the following 30 equations, which consider the rod length and sensor length:

$$\begin{bmatrix} |\overrightarrow{N_0 N_1}| - L_0 \\ |\overrightarrow{N_1 N_2}| - L_1 \\ |\overrightarrow{N_0 N_2}| - L_2 \end{bmatrix} = \mathbf{0} \tag{5}$$

Considering the length of the six struts ($L_r$), we can list the six equations constrained by the rigid bars, which cannot change in length:

$$\begin{bmatrix} |\overrightarrow{N_0N_3}| - L_r \\ |\overrightarrow{N_4N_5}| - L_r \\ |\overrightarrow{N_1N_6}| - L_r \\ |\overrightarrow{N_7N_8}| - L_r \\ |\overrightarrow{N_2N_9}| - L_r \\ |\overrightarrow{N_{10}N_{11}}| - L_r \end{bmatrix} = \mathbf{0} \qquad (6)$$

Considering the lengths of the other sensors, we can determine the positions of the remaining nodes:

$$\begin{bmatrix} |\overrightarrow{N_0N_7}| - L_3 \\ |\overrightarrow{N_0N_9}| - L_4 \\ \vdots \\ |\overrightarrow{N_8N_9}| - L_{22} \\ |\overrightarrow{N_8N_{11}}| - L_{23} \end{bmatrix} = \mathbf{0} \qquad (7)$$

Here, we adopted the least squares method to solve the equations above by starting with an initial guess for the values. Considering the dynamic condition, the resistance will change over time, thus, the position will update after getting the new value. The node positions in subsequent iterations are calculated from the prior iteration. In this case, the tensegrity will deform in the computer as the external force is exerted on the tensegrity, which can realize the shape recognition.

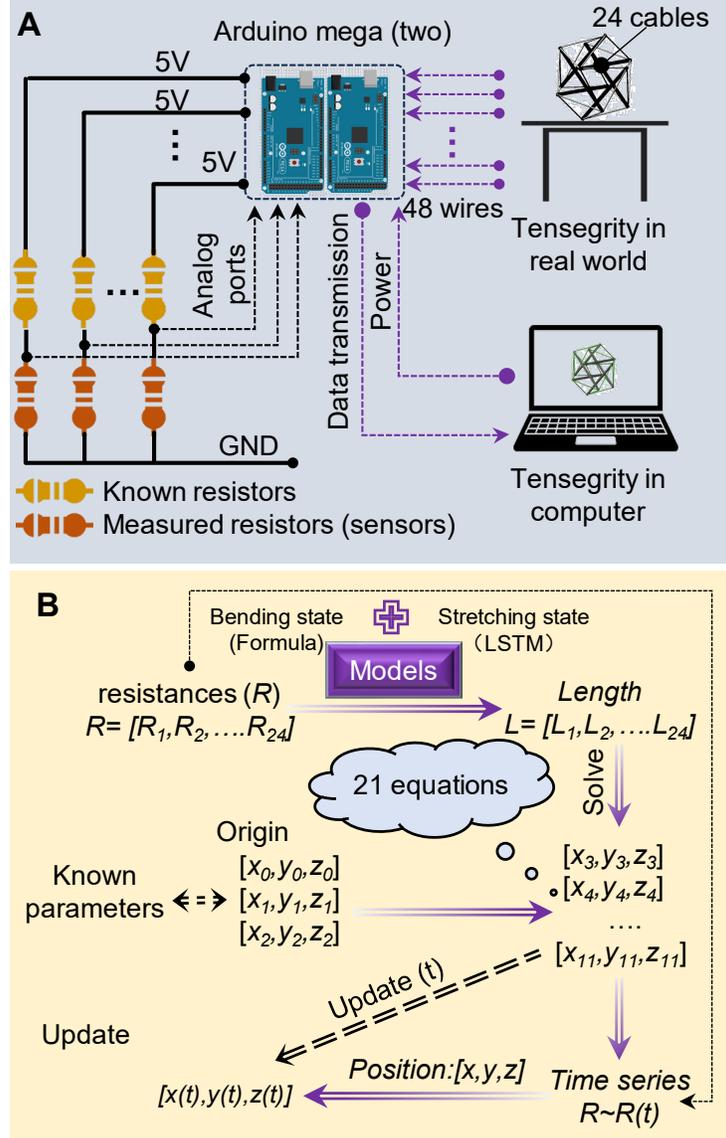

Fig. 2 Overview of intelligent tensegrity structure in real world and virtual world. **A:** Data acquisition module based on two sets of arduion mega. **B:** Shape recognition of tensegrity using the multimodal sensory system.

### III. CHARACTERISTICS AND DATA-DRIVEN MODELING OF MULTIMODAL STRAIN SENOR

We investigated the characteristics of a multimodal strain sensor made of diene rubber in terms of two states: bending and stretching. For the bending state, we used polynomial equations to fit the curve. For the stretching state, we utilized a LSTM network to build up the model.

*A. Bending strain sensor*

The sensors of the tensegrity structure have two states: a bending state when compressive strain exists and a stretching state when tensile strain exists. Here, we focused on the physical deformation properties, response time, and operational consistency under repetitive use of a flexible, durable strain sensor. Fig. 3**A** shows a flexible sensor made of diene rubber in both bending and normal states. In Fig. 3**B**, the sensor's resistance changes when it is deformed. Assuming the initial length and resistance values are $L_0$ and $R_0$, respectively, their values will become $L_1$ and $R_1$ as they are compressed. The resistance change can then be described as:

$$\Delta R/R = (R_1 - R_0)/R_0 \qquad (8)$$

The relationship between strain and resistance changes over time (Fig. 3**C**), indicating that as the strain increases, the resistance change also increases. We also tested the influence of bending rates (mm/min) on the performance of these sensors. The different colors represent different rates, suggesting that the sensor's response is consistent across these rates (Fig. 3**D**). Therefore, we fitted the curve using the following equation to express this relationship ($R^2$(coefficient of determination) = 0.9999):

$$y = -4.7589x^5 - 16.521x^4 - 20.239x^3 - 9.9675x^2 - 0.5464x - 0.0016 \quad (9)$$

The response times of the sensors were investigated, revealing two specific response times of 286 ms and 544 ms, respectively, highlighting their quick adaptability to strain changes (Fig. 3**E**). The durability of the sensor was tested by showing the change in resistance over 50 bending cycles. The consistent oscillation pattern without significant degradation indicates good durability (Fig. 3**F**). All experiments were conducted using the self-developed electromechanical tensile test equipment and an LCR meter (Hioki-3536, Hioki, Japan) [27].

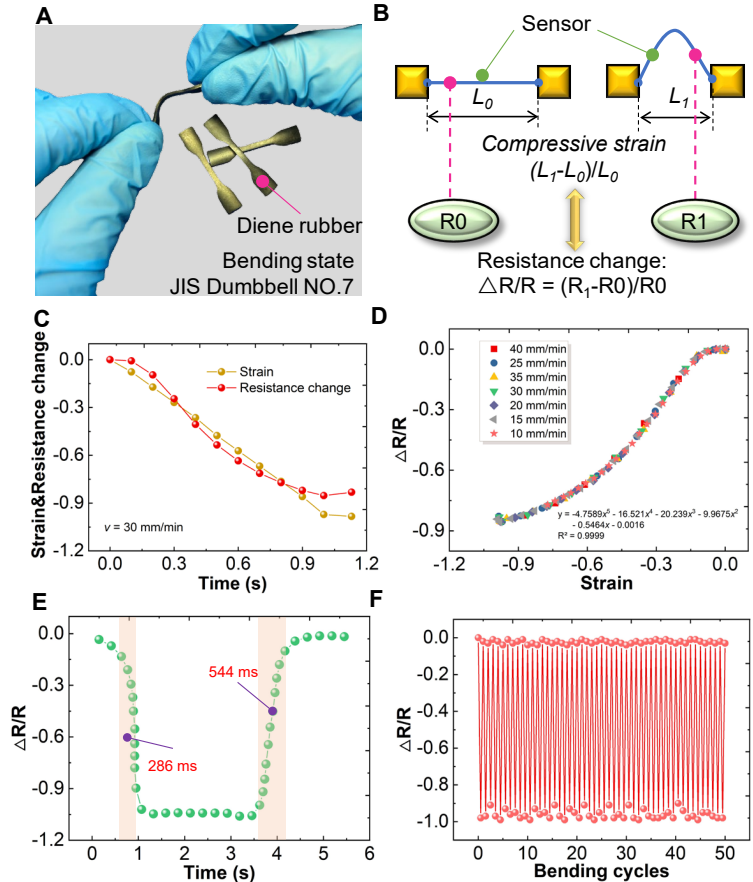

Fig. 3 Characteristics of bending state of the strain sensor. **A:** Flexible diene rubber sensor demonstrated in a bending state to show material pliability. **B:** Schematic of the strain sensor operation, depicting resistance change upon compressive strain. **C:** Correlation between strain and resistance change over time. **D:** Resistance response curves to different strains at varying rates, showing sensor consistency. **E:** Response time of the sensor at the bending state. **F:** Durability test of the sensor across 50 bending cycles.

*B. Stretching strain sensor*

We studied the performance of the strain sensor under various tests related to its stretchability and durability. Fig. 4**A** shows the stretching state of the sensors. We measured the resistance over 50 stretching cycles at a constant velocity of 0.5 mm/min, indicating the sensor's stability under cyclic loading (Fig. 4**B**). The tensile stress-strain curves, which demonstrate the mechanical properties of the material over repeated strain cycles, were obtained from the first (C1) to the fiftieth (C50) cycle (Fig. 4**C**). The sensor's sensitivity to deformation is shown in Fig. 4**D**, where the change in resistance (ΔR/R) is plotted against the strain for different gauge factors (GF1, GF2, GF3). Here, we focused on GF2. Fig. 4**E** shows the stability of the gauge factor (GF2) over numerous stretching cycles, demonstrating the sensor's nonlinear properties. Additionally, GF2 values vary across a wide range of strain levels, indicating the sensor's versatile response to various degrees of deformation (Fig. 4**F**). Moreover, we studied the impact of different stretching velocities on the resistance change, showing that the sensor's response is highly dependent on the stretching rate (Fig. 4**G**). The noise level of our sensor was also investigated, ranging between -0.23 and 0.13 (Fig. 4**H**). The response times, which are 281 ms and 926 ms, respectively (Fig. 4**I**).

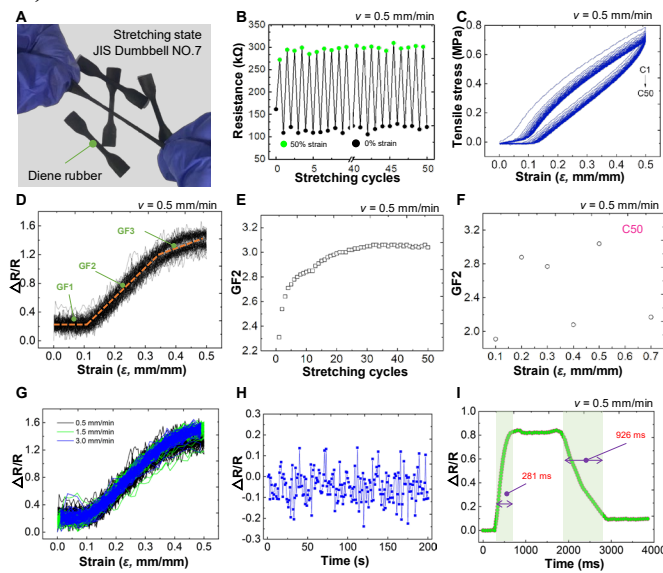

Fig. 4 Characteristics of stretching state of the strain sensor. **A:** Stretching demonstration. **B:** Resistance versus stretching cycles. **C:** Tensile stress-strain relationship. **D:** Resistance change and gauge factor versus strain. **E:** Gauge factor (GF2) stability. **F:** Gauge factor versatility. **G:** Resistance change versus strain under different stretching speeds. **H:** Noise levels of the sensor. **I:** Response time of the sensor in the stretching state.

## C. Deep learning assistive modeling

We used the deep learning method (Long Short-Term Memory) to construct the model for the multimodal strain sensors [28][29]. Fig. 5 illustrates various aspects of a machine learning model's development and evaluation process, specifically for time series data related to strain sensing. We prepared training and testing data with strain and resistance changes over time for different speeds. The data can be segmented into different velocity speeds for model training and testing (Fig. 5**A**). A LSTM cell is a type of recurrent neural network (RNN) used for time series analysis (Fig. 5**B**). The LSTM's components include input, forget, and output gates, as well as the cell state and hidden state dynamics. The forget gate outputs $\mathbf{f}_t$ (0~1), which determines how much of the information is discarded based on the cell state ($\mathbf{c}_{t-1}$):

$$\mathbf{f}_t = \sigma(\mathbf{W}_f[\mathbf{X}_t, \mathbf{h}_{t-1}] + \mathbf{b}_f) \qquad (10)$$

where $\sigma, \mathbf{W}_f, \mathbf{X}_t, \mathbf{h}_{t-1}$ and $\mathbf{b}_f$ are sigmoid function, weight matrix for the forget gate, short-term memory, observation vectors and bias term for the forget gate, respectively. The input gate $\mathbf{i}_t$ outputs which new information is to be stored in the cell (long-term memory) as:

$$\mathbf{i}_t = \sigma(\mathbf{W}_i[\mathbf{X}_t, \mathbf{h}_{t-1}] + \mathbf{b}_i) \qquad (11)$$

where $\mathbf{W}_i$ and $\mathbf{b}_i$ are weight matrix for the input gate, bias term for the input gate respectively. Also, the input gate decides how much of the new information will be stored in the cell state, creating new candidate values ($\tilde{\mathbf{c}}_t$) by combining the input and the previous hidden state:

$$\tilde{\mathbf{c}}_t = \tanh(\mathbf{W}_h[\mathbf{X}_t, \mathbf{h}_{t-1}]) \qquad (12)$$

where $\mathbf{W}_h$ is weight matrix for the input gate applied to the candidate cell state. The cell state $\mathbf{c}_t$ at time $t$ is updated forgetting the irrelevant parts of the previous cell state ($\mathbf{c}_{t-1}$, multiplication with the forget gate $\mathbf{f}_t$) and adding new candidate values ($\tilde{\mathbf{c}}_t$, multiplication with the forget gate $\mathbf{i}_t$).

$$c_t = \mathbf{f}_t \odot c_{t-1} + i_t \odot \tilde{c}_t \tag{13}$$

In the output gate ($o_t$), the amount of the cell state's information that is output as the hidden state is decided. The output gate uses the previous cell state and decides the hidden state:

$$o_t = \sigma(\mathbf{W}_o[X_t, h_{t-1}] + \mathbf{b}_o) \tag{14}$$
$$h_t = o_t \odot \tanh(c_t) \tag{15}$$

where $\mathbf{W}_o$, $h_t$ and $\mathbf{b}_o$ are weight matrix for the output gate, new hidden state at time $t$, bias term for the output gate respectively.

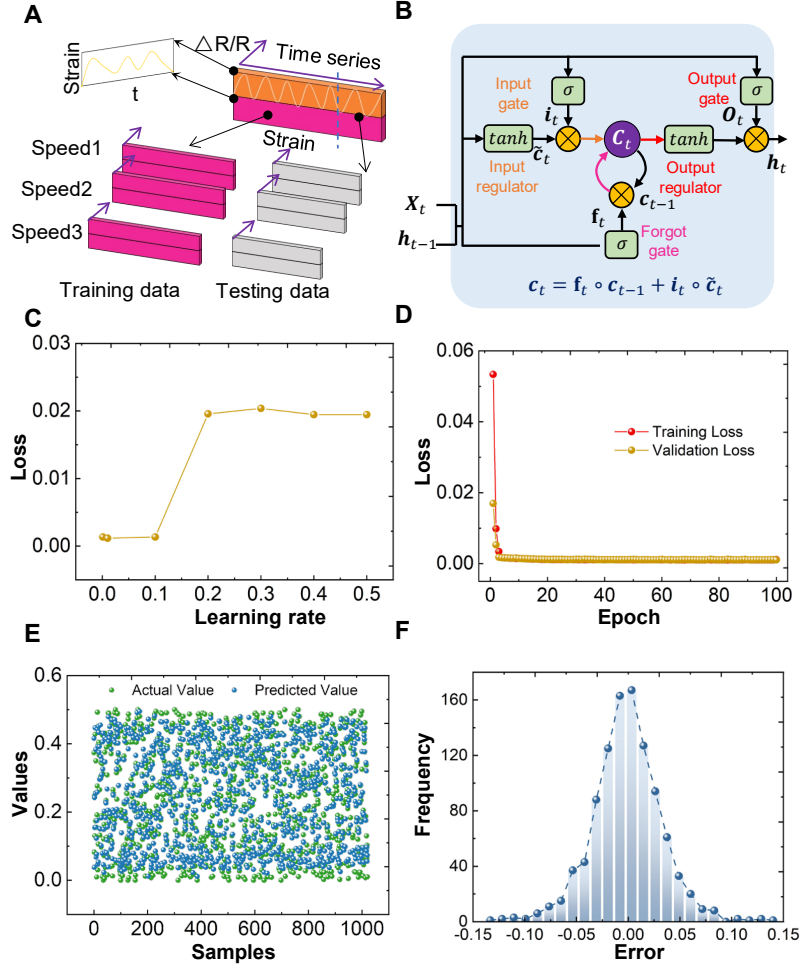

Fig. 5 Deep learning assistive learning model. **A:** Data preparation. **B:** Overall view of LSTM block processing. **C:** Learning rate optimization. **D:** Training process; **E:** Model predictions versus actual data relative to samples. **F:** Error distribution.

We optimized the loss (mean squared error) in terms of learning rate (Fig. 5**C**). The results show that the model loss is minimized when the optimal learning rate is in the range of 0 to 0.1. After the learning rate reaches 0.1, the loss increases. Therefore, we chose a learning rate of 0.1 as the model value. We also investigated the training and validation loss as a function of the number of epochs, which indicates the convergence of the model, with both training and validation loss decreasing and stabilizing over time (Fig. 5**D**). Moreover, the actual and predicted values generated by the model were compared (Fig. 5**E**), showing that our model can accurately estimate the values over a set of samples. Additionally, we researched the relationship between frequency and error (Fig. 5**F**). The histogram with a fitted curve of the prediction errors indicates that the error values are close to 0. Finally, we packaged the trained model for subsequent shape recognition and monitoring applications.

## IV. SHAPE RECOGNITION OF TENSEGRITY

The multimodal strain sensors of the tensegrity can be used to reconstruct the state of the robot through a combination of curve-fitting equations and LSTM modeling methods. To validate the model, we fabricated a 6-strut tensegrity structure (Fig. 6**A**). The struts are wooden rods with a diameter of 6 mm, and their ends are fixed by 9-mm diameter hollow cylinders for easy connection to the tendons. The tendons are made of long, flat, stretched strips of silicone rubber with a thickness of 2 mm, functioning as sensors. The tips of the tensegrity, referred to as caps, are produced using 3D printing technology (3D printer: Markforged® Mark-Two). Other components, including microcomputers, wires, and struts, were purchased.

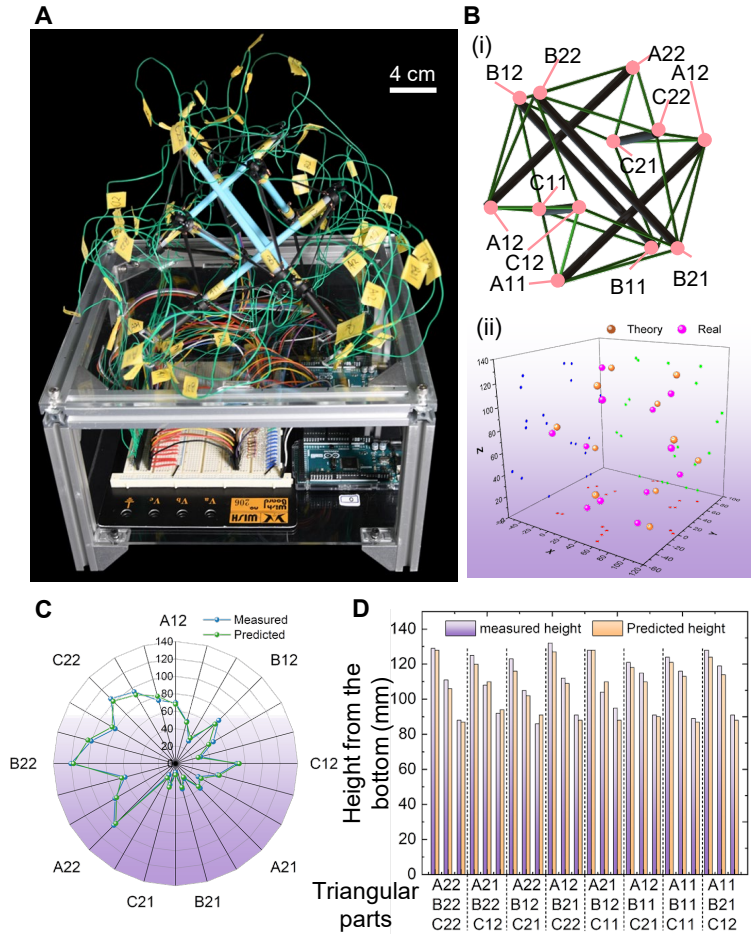

Fig. 6 Self-shape recognition of tensegrity. **A:** Experimental setup of multimodal sensor-equipped tensegrity structure. **B:** Nodes and data comparison between theoretical predictions and actual measurements. **C:** Height difference of the nodes between measured and estimated data. **D:** Height analysis comparing measured and predicted values for tensegrity faces.

We labeled the nodes of the tensegrity and compared the positions of each node based on theoretical calculations and actual measured data (Fig. 6**B**). The differences observed are attributed to fabrication skills. To determine the accuracy of the model, we compared the motion capture data with the points estimated using the multimodal strain sensors and the node position estimation algorithm. In our setup, the nodes A11, B11, and C11 are constrained and serve as known reference points. We simultaneously measured the position of the nodes as estimated by the state reconstruction model and the position (mainly height change (Z axis)) of the measured nodes as a human force was exerted on them, as shown in Fig. 6**C**. The root mean squared error (RMSE) of the data points (nodes) was calculated to be 21.2 mm. Moreover, we compared the measured heights against the predicted heights for various faces of the tensegrity (Fig. 6**D**). The RMSE for the estimated and measured heights of the planes is 35.8 mm. The

RMSE for the entire system is 39.4 mm. Fig. 7 shows state change of the tensegrity by applying forces to three nodes of the structure (A22, B22, and C22) and displays the reconstructed state using data from the multimodal sensors, calculated by the polynomial equations and the LSTM model. We also obtained the length changes of 24 sensors using a multi-line chart over 30 seconds (Fig. 8). Video **S1** shows the real-time deformation and state reconstruction of the tensegrity as the positions and quantities of nodes change.

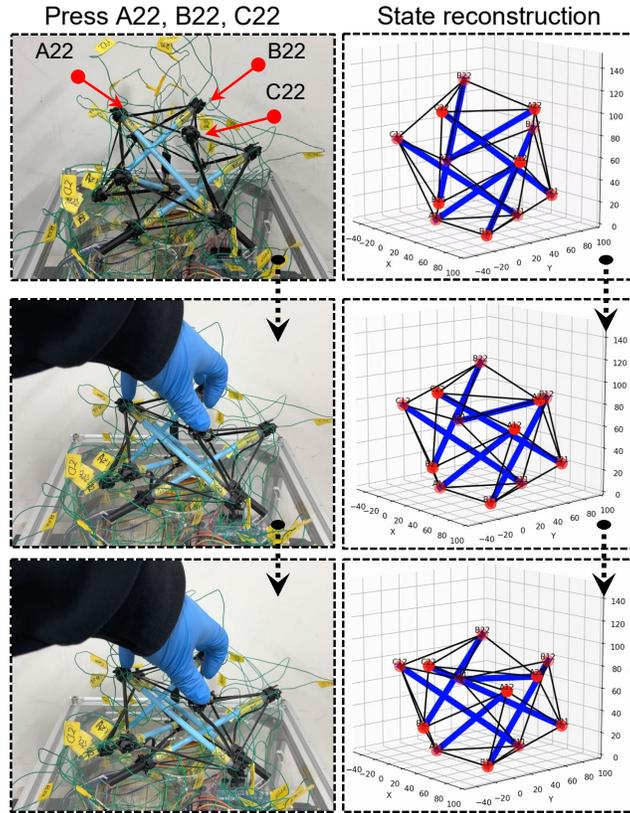

Fig. 7 Sequence of manual operation of the tensegrity with applying forces exerting at the nodes A22, B22, and C22. Visualization of state reconstruction derived from sensor data.

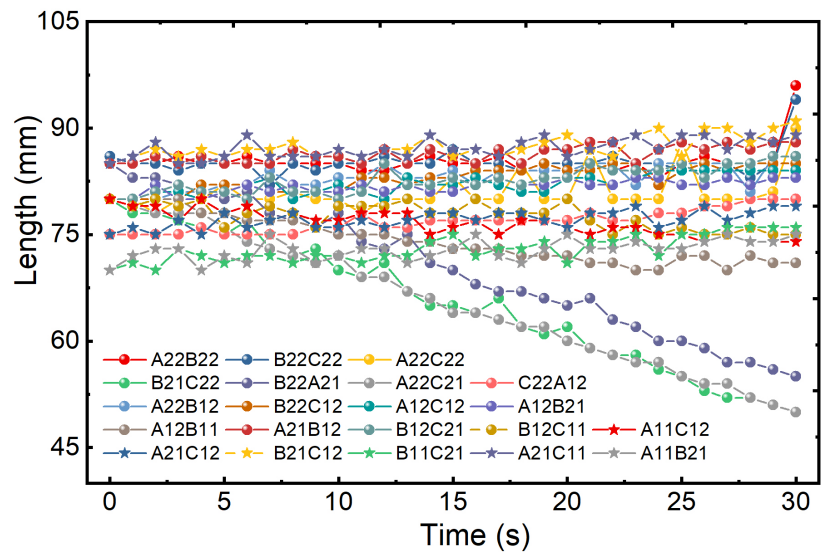

Fig. 8 Length changes of 24 sensors (tendons) over 30 s during the manual operation.

## V. Conclusion

In summary, this study presents an intelligent tensegrity-based system designed for exploring uneven and unpredictable environments. The integration of 24 soft multimodal sensors, replacing traditional tendons, allows for enhanced environmental interaction. The sensors have two modes: bending and stretching. In the bending state, their performance was curve-fitted using an empirical polynomial formula. In the stretching state, a well-trained LSTM model with 99% accuracy was developed to estimate the correlation between length and resistance. By combining these two models, the tensegrity equipped with a data acquisition module can achieve self-shape recognition, with RMSEs of 21.2 mm for nodes measurements and 35.8 mm for surface measurements, 39.4 mm for entire system, respectively. In future work, we will test the stability of the system under different environmental conditions, particularly varying temperatures and humidity levels. Table I lists the comparisons among tensegrities integrated with sensors. J. Bruce et al. used external sensors for the 6-strut tensegrity (9 kg) to achieve an RMSE of approximately 45.6 mm in terms of node state estimation. J. W. Booth et al. utilized internal sensors to predict the tensegrity's state with RMSE values of approximately 32.3 mm (node) and 45.8 mm (system). This comparison indicates that internal tensegrity structures generally tend to show a more consistent and possibly lower RMSE compared to external ones, although results vary significantly depending on the specific setup and parameters used in each study. In our study, we achieved RMSE values of 21.2 mm (node) and 39.4 mm (system), respectively, indicating high prediction accuracy for our system.

TABLE I
COMPARISON OF TENSEGRITIES' PERFORMANCE INTEGRATED WITH SENSORS

| Ref. | Tensegrity | Sensors | RMSE of shape recognition | |
|---|---|---|---|---|
| | | | Node (Best) | System |
| [21] | 6-strut | External | 9 kg | ~45.6 mm |
| [22] | | Internal | 2.55 kg | ~32.3 mm |
| [23] | 3-strut | Internal | —— | |
| [24] | 6-strut | External | 0.2 kg | —— |
| This study | | Internal | 0.4 kg | 21.2 mm |

ACKNOWLEDGMENT

We thank Prof. Shingo Maeda and Dr. Ardi Wiranata for their assistance in assembling the test rig.